# Hardware based Scale- and Rotation-Invariant Feature Extraction: A Retrospective Analysis and Future Directions

Shoaib Ehsan, Adrian F. Clark and Klaus D. McDonald-Maier
School of Computer Science & Electronic Engineering,
University of Essex,
Colchester, UK
sehsan@essex.ac.uk, alien@essex.ac.uk, kdm@essex.ac.uk

*Abstract*—Computer Vision techniques represent a class of algorithms that are highly computation and data intensive in nature. Generally, performance of these algorithms in terms of execution speed on desktop computers is far from real-time. Since real-time performance is desirable in many applications, special-purpose hardware is required in most cases to achieve this goal. Scale- and rotation-invariant local feature extraction is a low level computer vision task with very high computational complexity. The state-of-the-art algorithms that currently exist in this domain, like SIFT and SURF, suffer from slow execution speeds and at best can only achieve rates of 2-3 Hz on modern desktop computers. Hardware-based scale- and rotation-invariant local feature extraction is an emerging trend enabling real-time performance for these computationally complex algorithms. This paper takes a retrospective look at the advances made so far in this field, discusses the hardware design strategies employed and results achieved, identifies current research gaps and suggests future research directions.

## I. INTRODUCTION

"A picture is worth a thousand words" [1, 2 and 3] – this Chinese proverb illustrates concisely the significance of the utilization of rich information contained in an image. Image Processing and computer vision techniques represent a class of algorithms that focus on obtaining meaningful information from an image or sequence of images to achieve some high level objective like object recognition, target tracking and image matching. These algorithms are generally computation-intensive in nature and mostly implemented in software. Also, the data size involved is large, normally employing compression in storage and transmission.

Integral image calculation is a classic example that illustrates the computational complexity of these algorithms. An *integral image* is an intermediate representation of an image that allows fast calculation of rectangular features [4] and is particularly useful for multi-scale computer vision algorithms [5]; although comparatively new in the image processing domain, it has long been used in texture-mapping in computer graphics. Calculation of an integral image is a low-level computer vision task and only involves simple addition operations. However, for an input image of size M x N pixels, ¼ $M^2N^2$ additions are required to calculate integral image representation [6]. For example, 1,866,240,000 addition operations are required to calculate the integral image for an input image size of 360 x 240 pixels. The huge number of addition operations for a medium resolution image like this indicates the level of computational complexity in computer vision algorithms.

Real-time performance is desirable in many computer vision applications, such as target tracking and mobile robot navigation. However, with the high computational complexity of these algorithms, it is difficult to achieve the real-time goal with software-only implementation. Modern desktop computers that employ multiple processors clocking at GHz frequencies are, surprisingly, not generally well-suited to computation intensive, real-time computer vision algorithms due to the limited image processing capabilities of underlying hardware architecture. General-Purpose computation on Graphics Processing Units (GPGPU) is an emerging trend that utilizes high memory bandwidth and huge computational resources of graphics hardware to speed up many applications including image processing and video processing [7]. Graphics Processors, however, have very high power consumption (usually tens of watts) that makes them unsuitable for embedded vision applications with restricted power resources. Such applications usually employ commercial off-the-shelf (COTS) embedded computers that don't guarantee real-time performance as they run at much lower clock frequencies and have restricted computational and power resources as compared to graphics processors. Lack of computer architectures capable of processing image and video data is thus, a major hurdle in real-time vision processing.

In order to achieve real-time performance for computer vision applications, especially on embedded processors, the inherent parallelism found in this class of algorithms can be used to great advantage. This implies the design of special-purpose parallel hardware structures capable of real-time operation for specific algorithms and applications. In this paper, focus is on parallel hardware architectures for scale- and rotation-invariant local feature extraction, which serves as the primary stage of many computer vision applications. Scale and rotation invariance are desirable attributes for a feature extraction algorithm but bring with them significantly increased computational complexity. This paper provides a retrospective analysis, encompassing the design strategies employed and results achieved, for hardware architectures proposed so far in this area. Future directions for bridging the research gaps in this emerging field are presented at the end.

## II. SCALE AND ROTATION INVARIANT LOCAL FEATURE EXTRACTION

This section provides a brief review of state-of-the-art algorithms for scale- and rotation-invariant local feature extraction; an exhaustive survey of these algorithms can be found in [8]. Feature extraction, generally consisting of detection and description phases, falls in the category of low-level computer vision tasks. Corners, blobs and junctions are some generic local features that are usually of interest; however, the exact type of feature that needs to be extracted from any input image depends upon specific application.

Although local features have been around since the mid-1950s [9], there has been a trend towards scale- and rotation-invariant local features for solving wide variety of problems over the last few years, ranging from wide baseline stereo matching to the recognition of object classes [8]. Scale invariance and rotation invariance ideally imply that same image features can still be extracted if the input image is scaled up or down by any scale factor and rotated by any angle. This property is especially valuable if the vision system involves the analysis of moving imagery.

A number of computer vision algorithms have been proposed for extraction of scale and rotation invariant local features in the last decade or so. Some well-known examples are: Scale Invariant Feature Transform (SIFT) [10], Speeded-Up Robust Features (SURF) [11], Harris-Laplace/Affine and Hessian-Laplace/Affine feature detectors [12]. Of all the techniques presented so far, SIFT is widely regarded as the most popular algorithm in this domain. The SIFT algorithm employs a Difference of Gaussian (DoG) detector that has high repeatability; and its feature descriptor is highly distinctive and robust to changes in illumination, noise and minor view point changes. The main draw back of SIFT from the computational point of view is the high dimensionality of its descriptor, as that slows down the subsequent feature matching process. Another popular algorithm, focused more on execution speed of scale- and rotation-invariant feature extraction, is SURF. This is largely inspired by the SIFT algorithm but emphasizes performing fast detection, description and image matching.

The performance of these state-of-the-art algorithms on modern desktop computers is far from real-time due to the high level of computational complexity involved. In order to emphasize the computational complexity, execution times for software-only implementations of some popular scale- and rotation-invariant feature extraction schemes are given in Table I. These clearly demonstrate the inability of desktop computers to run these algorithms in real-time. For example, detection and description of 1529 interest points using the original software implementation of SURF [13] for first image (800 x 640 pixels) of the Graffiti data set provided by [14] takes about 610 ms on a standard Pentium-IV PC running at 3 GHz [11]. Special-purpose hardware architectures exploiting inherent parallelism of these algorithms are therefore required in order to achieve significant speed gain.

## III. HARDWARE BASED EXTRACTION

In the last few years, researchers have started to focus on hardware-based systems for real-time extraction of scale- and rotation-invariant image features. Hardware-based feature extraction is an emerging area which has made significant advances in a short period. Although there are a number of algorithms that can be targeted for hardware implementation, it is interesting to note that most of the work done in this area so far has been dominated by SIFT, and focused on designing an optimized hardware architecture for this technique. Other competing algorithms, such as SURF, have not yet been explored for hardware implementation – which indicates a research gap in this budding area. This section presents a survey of the technological advances made in this field over the last few years and discuses the results achieved.

TABLE I. PERFORMANCE OF SOME STATE-OF-THE-ART SCALE AND ROTATION INVARIANT FEATURE EXTRACTION ALGORITHMS ON MODERN DESKTOP COMPUTERS

| Algorithm | Computation Time with Platform description |
|---|---|
| SIFT | 600 ms for detection and description of interest points for an image size of 640 x 480 on Pentium III running at 700 MHz [15] |
| SURF | 610 ms for detection and description of 1529 interest points for an image size of 800 x 640 on Pentium IV running at 3GHz [11] |
| Harris-Laplace | 7 sec for detection of 1438 interest points for an image size of 800 x 640 on Pentium II running at 500 MHz [12] |
| Hessian-Laplace | 700 ms for detection of 1979 interest points for an image size of 800 x 640 on Pentium IV running at 3GHz [11] |
| Harris-Affine | 36 sec for detection of 1123 interest points for an image size of 800 x 640 on Pentium II running at 500 MHz [12, 16] |
| Hessian-Affine | 2.73 sec for detection of 1649 regions for an image size of 800 x 640 on Pentium IV Linux PC running at 2 GHz [17] |
| Maximally Stable Extremal Regions (MSER) | 140 ms for detection of regions for an image size of 530 x 350 on a Linux PC with the Athlon XP 1600+ Processor [18] |
| Salient Regions | 33 min and 33.89 sec for detection of 513 regions for an image size of 800 x 640 on Pentium IV Linux PC running at 2 GHz [17] |
| Edge-based Regions | 2 min and 44.59 sec for detection of 1265 regions for an image size of 800 x 640 on Pentium IV Linux PC running at 2 GHz [17] |
| Intensity-based Regions | 10.82 sec for detection of 679 regions for an image size of 800 x 640 on Pentium IV Linux PC running at 2 GHz [17] |

The work presented in [15] is considered ground breaking in the area of hardware-based scale- and rotation-invariant feature extraction. An FPGA based, fixed-point implementation of SIFT algorithm was targeted to achieve speed gain over software implementations. As a first step, a floating-point software implementation of the SIFT algorithm was converted to fixed-point, and modifications were made to routines so as to make them efficient for hardware implementation. Instead of using low-level hardware description languages like VHDL and Verilog, a high level tool known as System Generator was employed for major part of this particular hardware implementation. System Generator allows modeling and designing of signal processing systems targeted for Xilinx FPGAs in the MATLAB-SIMULINK environment, using the Xilinx Blockset that contains signal processing functions like FIR filter and Viterbi decoder [19]. In this work, VHDL was used only for implementing low-level processes like DMA transfers and memory accesses, to make them more efficient. The final bit file for the FPGA was generated using Xilinx ISE. The Virtex-II Xilinx FPGA-based design reduced execution time of SIFT to 60 ms for an image size of 640 x 480 pixels, compared to 600 ms required on a Pentium-III 700 MHz processor. In [20], it is reported that this FPGA-based design is capable of computing SIFT features at a rate of 7 Hz for an image size of 1024 x 768. No other specific details, such as FPGA resource utilization or power consumption, are available about the designed hardware. Accuracy of the fixed-point hardware relative to the floating-point software implementation is also not discussed.

A partial hardware implementation of the SIFT algorithm is reported in [21] for online stereo calibration. Only two main components of the SIFT algorithm, i.e., Gaussian pyramid and Sobel filter, were implemented in Virtex-II Xilinx FPGA using VHDL, whereas the remaining ones were executed in software on a host computer. A pipelined hardware architecture clocking at 54 MHz was designed for implementing the Gaussian pyramid in a way that allowed feature extraction to start before the image was fully digitized. A Sobel operator was also implemented in FPGA, instead of calculating edge gradient and orientation using finite differences. This architecture was capable of operating at 60 frames per second and reduced by 50–70% the time for feature extraction. No information is provided regarding input image resolution, FPGA resource utilization and power consumption of the designed hardware.

An innovative pipelined hardware architecture for Harris-Affine feature detector was presented in [16, 22] and was claimed to be the first attempt at implementation of a complex iterative algorithm in reprogrammable hardware. This fixed-point implementation was unique in a sense that it employed multiple FPGAs for extraction of scale- and rotation-invariant features. The coding was done in VHDL and was compiled using the Quartus-II software provided by Altera [23]. The computations were distributed among four Altera Stratix S80 FPGAs that were able to process standard video (640 x 480 pixels) at 30 frames per second. As a reference, a Stratix S80 FPGA consists of 79040 logic elements, 176 9-bit DSP elements and 7427520 bits of memory. The resource utilization per FPGA for this design is given in Table II and it shows that the available resources were not fully-utilized. This hardware architecture achieved a speed gain of 90-9000 times over an equivalent software implementation of the Harris-Affine feature detector, depending upon the language of implementation and the computing platform. Although comprehensive detail about the designed architecture was included in [16], there was no discussion about the power consumption.

Another FPGA-based partial implementation of the SIFT algorithm is reported in [24, 25]. A hardware-software co-design strategy was preferred over pure hardware implementation; the hardware–software partitioning was done in such a way that the detection phase of the algorithm was implemented in hardware whereas the description phase was targeted to run in software on a MicroBlaze processor. This architecture was realized on a Xilinx XUP-Virtex-II Pro board but was only capable of processing one octave for the SIFT algorithm. With MicroBlaze running at 100 MHz, it was claimed that this architecture required 0.8 ms for detection and description of key points for an image size of 320 x 240 pixels [24]. However, according to [25], 0.8 ms is required for an image size of 340 x 240 pixels.

An FPGA based implementation of the Maximally Stable Extremal Region (MSER) detector was described in [26]. The designed architecture was implemented on a Xilinx XC2VP100 FPGA and achieved performance of 54 frames per second for an image size of 320 x 240 pixels without using any off-chip memory. Important implementation results for the designed architecture are given in Table III.

In [27], a dedicated processor for SIFT-based object recognition was proposed for the first time. This processor was based on Visual Image Processing memory and Network-on-Chip. Ten SIMD processing elements were also integrated into this processor architecture for exploiting data- and task-level parallelism of the SIFT algorithm. An important feature of this architecture was its low-power consumption. For an input image size of 320 x 240, this dedicated processor was able to achieve 10.1–15.9 frames per second for SIFT feature extraction at 200 MHz. Implementation results for this processor are summarized in Table IV.

Finally, a parallel hardware architecture for SIFT is proposed in [28] which utilizes a hardware–software co-design strategy. Excepting descriptor computation, which runs in software on a NIOS-II soft core processor, all other steps of the SIFT algorithm are implemented in hardware. This is the most complete implementation of the SIFT algorithm to date and provides accurate results that are similar to software implementations. With a NIOS-II soft core processor running at 100 MHz, this architecture requires 33ms to extract SIFT features for an image size of 320 x 240 pixels; thus, it can achieve performance of up to 30 frames per second.

TABLE II. RESOURCE UTILIZATION PER FPGA FOR HARDWARE BASED HARRIS-AFFINE DETECTOR [16, 22]

| FPGA | Logic Elements | Memory Bits | DSP Elements |
|---|---|---|---|
| FPGA 0 | 38256 (48%) | 3527389 (47%) | 126 (72%) |
| FPGA 1 | 55028 (70%) | 5045677 (68%) | 112 (64%) |
| FPGA 2 | 37036 (47%) | 4159329 (56%) | 25 (14%) |
| FPGA 3 | 70159179 (89%) | 4437738 (60%) | 117 (66%) |

TABLE III. IMPLEMENTATION RESULTS FOR THE FPGA BASED MAXIMALLY STABLE EXTREMAL REGIONS (MSER) DETECTOR PRESENTED IN [26]

| Image Resolution | 320 x 240 pixels |
|---|---|
| LUTs | 9800 (11%) |
| Memory | 4.6 MBit (63%) |
| Execution Time | ~ 19 ms |
| Speed | 42 MHz |

TABLE IV. IMPLEMENTATION RESULTS FOR THE DEDICATED OBJECT RECOGNITION PROCESSOR PRESENTED IN [27]

| Technology | 0.18 µm 1-poly 6 Metal Standard CMOS |
|---|---|
| Chip Size | 7.7 mm x 5 mm |
| Clock Frequency | 400 MHz for NoC, 200 MHz for other parts |
| Gate Count (NAND2 Equiv.) | 838.8 K |
| On-Chip Memory | 34 Kbytes |
| Peak Power Consumption | 1.4 W at 1.8 V |

IV. FUTURE DIRECTIONS

The field of hardware-based scale- and rotation-invariant feature extraction is still in its infancy, despite making substantial advances in the last few years. Achievement of real-time performance has been the motivating factor in this area so far. However, it is important to identify current research gaps in this field and determine future directions and goals early. This section presents some prospective directions that can lead to fast maturity of this area.

A big research gap is the lack of low-power hardware architectures that are capable of real-time extraction of scale- and rotation-invariant features. The work presented in [27] is the sole effort to date that attempts to bridge this gap. Vision processing is vital for many embedded applications to achieve some high-level objective. However, battery-powered embedded systems generally have a restricted power budget and their operation time is dependent upon system power consumption. Running computation-intensive feature extraction algorithms results in fast depletion of power resources in embedded vision applications and might lead to an incomplete mission. Design of low-power hardware architectures capable of real-time feature extraction will not only solve the power consumption bottlenecks in existing embedded vision applications but will also pave way for exciting new applications. A classic example of this is small unmanned aerial vehicles (UAVs): they are generally equipped with still and/or video camera(s) that capture crucial image information but are unable to process it on-board due to strict power constraints. Low-power hardware architectures will reduce the dependence of UAVs on ground stations for vision processing tasks and thus will make them more autonomous. Another interesting application is a wireless sensor network. A wireless image sensor node capable of extracting scale- and rotation-invariant image features would only need to transmit the extracted features to other nodes, thus drastically reducing the amount of data transmitted across the network. It is, therefore, time to investigate feature extraction algorithms having relatively low computational complexity and acceptable level of accuracy for designing low-power hardware architectures. A potential candidate for mapping to low-power hardware seems to be the SURF algorithm as it exploits the integral image representation for the rapid calculation of box-type filters. Employment of the integral image representation allows the SURF algorithm to avoid multiplications and reduces the computational complexity to three simple addition operations for box-filter calculation. As mentioned above, there has been no effort to design any kind of special-purpose hardware for the SURF algorithm so far. It is, therefore, worthwhile to evaluate the potential of SURF for low-power hardware implementation in future research. The Linear Time Maximally Stable Extremal Regions algorithm [29] seems to be another prospective technique for low-power hardware implementation and may be explored in future.

Another research gap that needs to be bridged is the exploration of multi-processor architectures. These architectures can be used to great advantage, due to the inherent parallelism of scale- and rotation-invariant feature extraction algorithms. Multi-processor architectures would potentially permit real-time execution of these computation-intensive algorithms at relatively lower clock rates, and would also reduce over all power consumption. Investigation of multi-processor architectures for scale- and rotation-invariant feature extraction algorithms is therefore another potential direction.

Research efforts to date have been directed at designing hardware architectures that are optimized for some specific feature extraction algorithm. This actually limits the application domain of the designed hardware, as features of interest are not the same for all applications. For example, the Harris-Laplace detector is useful in applications where corners need to be detected but is not well-suited for applications requiring blob detection. Thus, instead of designing separate hardware architectures for each individual algorithm, it would really be valuable to design a generic hardware architecture capable of real-time performance for multiple feature extraction algorithms. The possibility of designing such generic architectures may be investigated as a next step forward in this area.

## V. CONCLUSIONS

Current state-of-the-art computer vision algorithms for scale- and rotation-invariant local feature extraction are highly computation-intensive in nature and their performance in terms of execution speed on modern desktop computers is far from real-time. There has been a recent trend towards hardware-based feature extraction to achieve performance gain. This paper has presented a survey of significant advances made in this field so far and identified current research gaps that need to be bridged. It has suggested that low-power and multi-processor architectures may be investigated in future research, as they have the potential to pave way for exciting new applications ranging from UAVs to wireless sensor networks. Another step forward in this emerging field will be to explore the possibility of developing generic hardware architectures that are capable of real-time performance for multiple feature extraction schemes. This would help to broaden the application spectrum of the designed hardware architectures.